\documentclass[]{article}

\pdfoutput=1

\usepackage[utf8]{inputenc}
\usepackage[english]{babel}

\usepackage{amsmath} %for theorems, definitions
\usepackage{bm} %for bold math symbols in definitions

\newtheorem{definition}{Definition}
\newtheorem{lemma}{Lemma}

\usepackage{algpseudocode, algorithm,algorithmicx}
\algtext*{EndFunction}% Remove "end function" text
\algtext*{EndFor}% Remove "end for" text
\algtext*{EndWhile}% Remove "end while" text
\algtext*{EndIf}% Remove "end if" text
\newcommand{\Continue}{\State \textbf{continue} }

\usepackage{graphicx} %for includegraphics

\usepackage[backend=bibtex,style=numeric]{biblatex}
\bibliography{bibliography.bib}
\title{Linear time DBSCAN for sorted 1D data and laser range scan segmentation}
\author{Bartosz Meglicki \\ Łukasiewicz Research Network \\ Industrial Research Institute for Automation and Measurements \\ PIAP \\ Warsaw, Poland}

\begin{document}

\maketitle

\begin{abstract}

This paper introduces new algorithm for line extraction from laser range data including methodology for efficient computation. The task is cast to series of one dimensional problems in various spaces. A fast and simple specialization of DBSCAN algorithm is proposed to solve one dimensional subproblems. Experiments suggest that the method is suitable for real-time applications, handles noise well and may be useful in practice.

\end{abstract}

\section{Introduction}

Density Based Spatial Clustering of Applications with Noise (DBSCAN) is an algorithm introduced in 1996 \cite{original} which received SIGKDD Test of Time Award \cite{award} in 2014. The award note praises the work for ability to find clusters of arbitrary shape, robustness to noise and support for large databases under reasonable conditions. Unlike other classic methods like k-means
\cite{forgy, bell}, DBSCAN does not require user to know the number of clusters a priori.

This paper focuses on efficient specialization of DBSCAN for 1D data and real-time application for 2D laser range scan segmentation. The benefits include the algorithm speed, ability to handle noise and no need for prior knowledge about number of clusters.

\subsection{Complexity Confusion}

There is much confusion about DBSCAN complexity in the literature. The original paper \cite{original} claimed average $O(NlogN)$ complexity under condition that $\varepsilon$-neighborhoods are small compared to the input size and appropriate data structures are used. Master thesis \cite{faster} pointed that the algorithm runs in pessimistic $O(N^2)$ time and subsequently improved versions by various authors either also run in $O(N^2)$ or do not calculate the same clustering as original DBSCAN \cite{original}. For unknown reasons the idea that DBSCAN algorithm runs in pessimistic $O(NlogN)$ time (not even the average!) was deeply rooted in literature, both journal articles and textbooks, as is pointed out in \cite{revisited} awarded as SIGMOD 2015 Best Paper. The same authors in \cite{revisited_long} go even further and point out that the notion of average complexity in original \cite{original} does not follow standard definition in computer science and question the expectation that $\varepsilon$-neighborhoods are small compared to input size.  

The work in \cite{faster} introduced 2D grid based algorithm that truly runs in pessimistic $O(N minPoints + NlogN)$ time, where $minPoints$ is
one of original DBSCAN algorithm parameters. In \cite{revisited_long} authors introduce a 2D grid solution that works in $O(N)$ for data presorted on both dimensions.

In this paper we focus on much simpler 1D algorithm that also runs in $O(N)$ time for presorted data. This solution is tailored specifically for application, line extraction from laser range data. Interestingly this 2D problem is cast to series of 1D clustering problems in various spaces.

\subsection{Preliminaries}

The DBSCAN algorithm resembles a classic Flood Fill algorithm used in  graphics programs as bucket tool. As opposed to Flood Fill, DBSCAN works in continuous domain painting dense regions with cluster identifiers. The notion of connectivity is not easy to define in continuous domain. DBSCAN introduces idea of core-points to spread information. Such points are required to have dense neighborhood. In this section we restate some of the definitions introduced in the original DBSCAN paper \cite{original} with simple examples in one dimension.

The $\varepsilon$-neighborhood of a point $p$ consists of all the points that lie within distance $\varepsilon$ from the point $p$.

\begin{definition}
	 Let $X$ be a set of points. The \textbf{\bm{$\varepsilon$}-neighborhood} of point p, denoted by $N_{\varepsilon}(p)$, is defined by $N_{\varepsilon}(p)=\{q \in X | distance(p,q) \leq \varepsilon \}$.   	
\end{definition}

\begin{figure}[h] 
	\centering
	\includegraphics{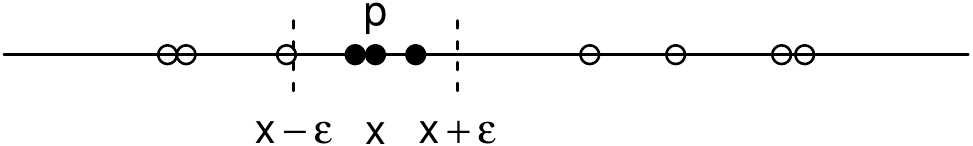}
	\caption{$\varepsilon$-neighborhood of a point (filled circles).}
\end{figure}

If $\varepsilon$-neighborhood of a point $p$ contains at least defined number of points the point $p$ is called a core point. 

\begin{definition}
	A point $p$ is a \textbf{core point} if $|N_\varepsilon(p)| \geq minPoints$. 	
\end{definition}

\begin{figure}[h] 
	\centering
	\includegraphics{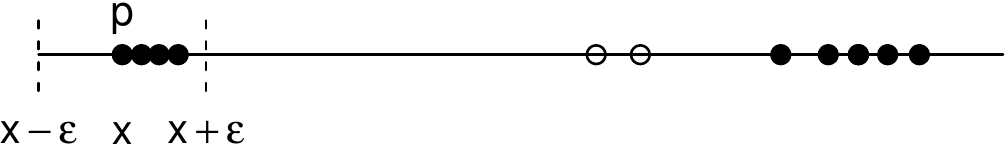}
	\caption{Core points (filled circles), $minPoints=4$, leftmost  core point $p$ is used to illustrates $\varepsilon$ value.}
\end{figure}

We will say that points close to the core points are directly density reachable from them.

\begin{definition}
	\label{def:directly_density_reachable}
	A point $p$ is \textbf{directly density reachable} from a point $q$ with respect to $\varepsilon$ and $minPoints$ if $p \in N_\varepsilon(q)$ and $q$ is a core point. 	
\end{definition}

\begin{figure}[h] 
	\centering
	\includegraphics{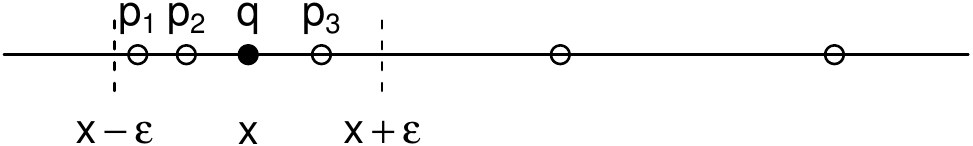}
	\caption{Points $p_1, p_2, p_3, q$ are directly density reachable from $q$ (filled circle), $minPoints=4$. The only core point $q$ is used to illustrate $\varepsilon$ value. Points $p_1, p_2, p_3$ are border points.}
\end{figure}

Points that are directly density reachable from some core point $q$ but don't have enough points in their neighborhood to be core points themselves are called border points.

\begin{definition}
	A point $p$ is a \textbf{border point} if $p$ is directly density reachable from some core point $q$ and $p$ is not a core point.
\end{definition}

If there is a sequence of points starting from $q$ and ending on $p$ where consecutive points are directly density reachable, we say that $p$ is density reachable from $q$. 

\begin{figure}[h] 
	\centering
	\includegraphics{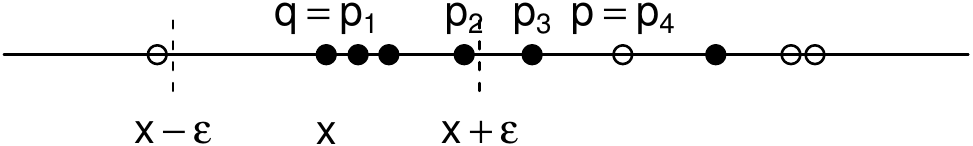}
	\caption{Point $p$ is density reachable from $q$ through e.g. $p_1, p_2, p_3, p_4$. Point $q$ is not density reachable from the point $p$. The core points are marked with filled circles, $minPoints=4$. The point $q$ is used to illustrate $\varepsilon$ value.}
\end{figure}

\begin{definition}
	\label{def:density_reachable}
	A point $p$ is \textbf{density reachable} from a point $q$ with respect to $\varepsilon$ and $minPoints$ if there is a chain of points $p_1, ..., p_n$ with $p_1=q$ and $p_n=p$ such that $p_{i+1}$ is directly density reachable from $p_i$. 	
\end{definition}

If points $p$ and $q$ are both density reachable from some point $o$, we say that $p$ and $q$ are density connected.

\begin{figure}[h] 
	\centering
	\includegraphics{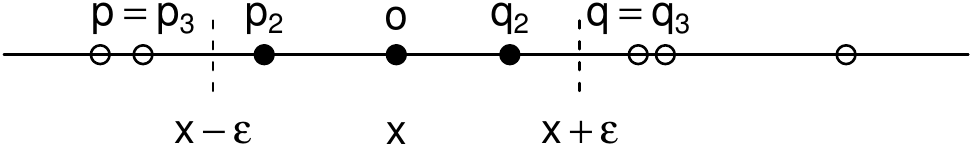}
	\caption{Point $p$ is density connected to $q$ through $o$. Point $p$ is density reachable from $o$ through chain $o, p_2, p_3$ and point $q$ is density reachable from $o$ through $o, q_2, q_3$. The core points are marked with filled circles, $minPoints=4$. The point $o$ is used to illustrate $\varepsilon$ value.}
\end{figure}

\begin{definition}
	A point $p$ is \textbf{density connected} to point $q$ with respect to $\varepsilon$ and $minPoints$ if there is a point $o$ such that both $p$ and $q$ are density reachable from $o$. 	
\end{definition}

A cluster is a set of density connected points maximal with respect to density reachable relation.  All the points density reachable from cluster point also belong to this cluster. 

\begin{definition}
	Let $X$ bet a set of points. A \textbf{cluster} $C$ with respect to $\varepsilon$ and $minPoints$ is a non empty subset of $X$ satisfying conditions:
	\begin{enumerate}
		\item $\forall p,q \in C$ : $p$ is density connected to $q$ with respect to $\varepsilon$ and $minPoints$
		\item $\forall p, q \in X$ : if $p \in C$ and $q$ is density reachable from $p$ with respect to $\varepsilon$ and $minPoints$, then $q \in C$
	\end{enumerate} 	
\end{definition}

The points that don't belong to any cluster are classified as noise. The definition is general enough to take into account clusters with respect to distinct $\varepsilon$ and $minPoints$ parameters.

\begin{definition}
	Let $C_1, ..., C_k$ be the clusters of the set of points $X$ with respect to $\varepsilon_i$ and $minPoints_i$, $i=1, ..., k$. We define \textbf{noise} as the set of points from $X$ that don't belong to any cluster $C_i$, i.e. $noise=\{p \in X | \forall i : p \notin C_i  \}$.
\end{definition}

\subsubsection{Border Points}

Note that by the above definitions a border point may belong to more than a single cluster. This happens if it is density reachable from some points that belong to distinct clusters. The problem is resolved in the original paper \cite{original} by assigning border points to the first found cluster they belong to. Some works \cite{border_noise} treat border points as noise instead.

\section{Clustering Alghorithm}

The original paper \cite{original} introduces two lemmas that simplify finding the clusters. 

\subsection{Finding the Clusters}

The first lemma states that for a given parameters $\varepsilon$ and $minPoints$ we can take an arbitrary core point $p$ from $X$ and the points that are density reachable from $p$ form a cluster.  

\begin{lemma}
	\label{lemma:core_to_cluster}
Let $p$ be a core point in $X$. Then the set $O=\{o|o \in X$ and $o$ is density reachable from $p$ with respect to $\varepsilon$ and $minPoints$\} is a cluster with respect to $\varepsilon$ and $minPoints$.
\end{lemma}

The second lemma states that any cluster with respect to $\varepsilon$ and $minPoints$ is uniquely determined by any of its core point $p$ and points density reachable from $p$.

\begin{lemma}
	\label{lemma:cluster_by_core}
	Let $C$ be a cluster with respect to $\varepsilon$ and $minPoints$ and let $p$ be any core point in $C$. Then $C$ equals to the set $O=\{o|o \in X$ and $o$ is density reachable from $p$ with respect to $\varepsilon$ and $minPoints\}$.

\end{lemma}

\subsection{Algorithm Overview}

By lemmas \ref{lemma:core_to_cluster} and \ref{lemma:cluster_by_core} we can start with arbitrary point $p$ in $X$ and points density reachable from $p$ will form a cluster. If $p$ is not a core point no points will be density reachable from it. In either case we can move to the next point in $X$ in the search for next cluster.

The problem is in finding the clusters efficiently. As we shall see, for the sorted 1D data, we can calculate $\varepsilon$-neighborhood of all the points in $X$ in $O(n)$ time. Using this information we can query the size of arbitrary point $\varepsilon$-neighborhood in $O(1)$. Finally we can move through points in $X$ expanding the clusters efficiently according to lemmas \ref{lemma:core_to_cluster} and \ref{lemma:cluster_by_core}.  

The DBSCAN1D algorithm \ref{alg:dbscan1d} takes as input the set of sorted input points $X$ and parameters $\varepsilon$, $minPoints$ for calculation of $\varepsilon$-neighborhood and clusters. In lines 2-6 the algorithm initializes cluster labels, identifier for next found cluster and list of found clusters.

In line 7 the algorithm calculates the upper $U$ and lower $L$ bound indexes for $\varepsilon$-neighborhoods of all the points in $X$.

In lines 8-16 the algorithm moves through the points in $X$ expanding the clusters if necessary and adding found clusters to the list. Finally the cluster list is returned in line 17.

\begin{algorithm}
	\caption{DBSCAN for 1D sorted data}
	\begin{algorithmic}[1]
		\Require{$X=[x_0,...,x_{N-1}]$ sorted input points, $x_0 \leq ... \leq x_{N-1}$} 
		\Require{$\varepsilon > 0$} 
		\Require{$minPoints > 0$}

		\Statex
		
		\Function{Dbscan1D}{$X,\varepsilon, minPts $}
		\State $C = [c_0, ..., c_{N-1}]$ 
		\State $clusterId \gets 1$
		\State $Clusters \gets \emptyset $

		\Statex
		\For{$i \gets 0$ to $N-1$}  
			\State{$c_i \gets NOT\_VISITED$}
		\EndFor	
		
		\Statex

		\State $(L, U) \gets$ \Call{CalculateNeighborhood}{$X, \varepsilon$}
		
		\Statex
		
		\For{$i \gets 0$ to $N-1$}  	
			\If {$c_i \neq NOT\_VISITED$}
				\Continue
			\EndIf
			
			\If {\Call{NeighborhoodSize}{$i, L, U$} $< minPts$}
				\State $c_i \gets NOISE$
			\Else
				\State $Cluster \gets$ \Call{ExpandCluster}{$X, i, L, U, C, clusterId, minPts$}
				\State $Clusters \gets Clusters \cup Cluster$ 
				\State $clusterId \gets clusterId + 1$
						
			\EndIf
			
		\EndFor
		\State \Return{$Clusters$}
		\EndFunction
	\end{algorithmic}
	\label{alg:dbscan1d}
\end{algorithm}

We now focus on efficient implementation of the algorithm subroutines, starting from $\varepsilon$-neighborhood calculation called in line 7.

\subsection{Calculating Neighborhood}

	\label{sub:calculating_neighborhood}

Let $X=[x_0,...,x_{N-1}]$ be the sorted table of input points. For arbitrary point $x_i$ and $\varepsilon$ value let $u_i$ be the inclusive upper bound index for $\varepsilon$-neighborhood of the point $x_i$. Formally $u_i$ is the largest $j$ such that $x_j \leq x_i + \varepsilon$ where $i \leq j < N$.

In a symmetric manner, let $l_i$ be the inclusive lower bound index for $\varepsilon$-neighborhood of point $x_i$. Formally $l_i$ is the smallest index $j$ such that $x_j \geq x_i - \varepsilon$ where $0 \leq j \leq i$.

By those definitions we have that the $\varepsilon$-neighborhood $N_\varepsilon(x_i)=\{x_{l_i}, ..., x_{u_i}\}$. Note that if we have two consecutive points $x_i$ and $x_{i+1}$ the upper bound index of $x_{i+1}$ is greater or equal the upper bound index of a previous point $x_i$, formally $u_{i+1} \geq u_i$.

\begin{figure}[h] 
	\centering
	\includegraphics[scale=0.8]{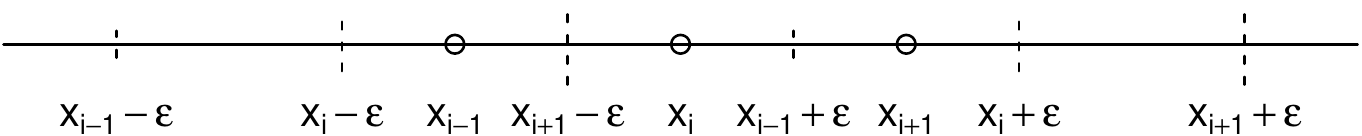}
	\caption{$\varepsilon$-neighborhoods range of consecutive points. From the ordering we have $x_{i+1} \geq x_i$ hence $x_{i+1} + \varepsilon \geq x_i + \varepsilon$ and finally $u_{i+1} \geq u_i $. The case for lower bounds is symmetric.}  
\end{figure}

In a symmetric manner if we have two consecutive points $x_{i-1}$ and $x_i$ their lower bounds are ordered accordingly, namely $l_{i-1} \leq l_i$.

Those simple observations are crucial for efficient computation of lower and upper bounds $L=[l_0,...,l_{N-1}]$ and $U=[u_0,...,u_{N-1}]$. For the upper bounds $U$, we will start with $x_0$ and iterate finding $u_0$. We then move to $x_1$ and from the inequality $u_{i+1} \geq u_i$ we can pickup the search for $u_1$ where we finished for $u_0$. By advancing this way to $u_{N-1}$, in a single pass through the points in $X$ we calculate all the upper bounds $U=[u_0,...,u_{N-1}]$.

The case for the lower bounds $L$ is symmetric. We start from $x_{N-1}$ and move down exploiting inequality $l_{i-1} \leq l_i$. By the time we arrive at $x_0$ we have calculated all the bounds $L=[l_0,...,l_{N-1}]$ in a single pass through the points in $X$.

The CalculateNeighborhood algorithm \ref{alg:calculate_neighborhood} implements those two searches. It takes as input the set of sorted input points $X$ and $\varepsilon$ parameter for calculating $\varepsilon$-neighborhood. In lines 2-5 the algorithm initializes the tables $U$ and $L$ and sets initial values for variables $u$ and $l$ that will keep the last found bounds.

Lines 6-9 implement the pass through points in $X$ calculating the table $U$. Symmetric lines 10-13 implement the search for table $L$. The result tables are returned in line 14.

The $O(N)$ complexity of lines 6-9 follows from the fact that the inner loop in line 7 always picks up the search where it was last finished and the search can only advance. The case for lines 10-13 is symmetric which warrants the $O(N)$ complexity of the whole algorithm.  
 
\begin{algorithm}
\caption{Calculating the neighborhood}
\begin{algorithmic}[1]
	\Require{$X=[x_0,...,x_{N-1}]$ sorted input points, $x_0 \leq ... \leq x_{N-1}$} 
	\Require{$\varepsilon > 0$}
 		
	\Statex
	\Function{CalculateNeighborhood}{$X, \varepsilon$}
	
	\State $U=[u_0, ..., u_{N-1}]$
	\State $L=[l_0, ..., l_{N-1}]$
	
	\State $u \gets 0$
	\State $l \gets N-1$
	\Statex
	
	\For {$i \gets 0$ to $N-1 $}
		\While {$u<N \land \lvert x_i-x_u \rvert \leq \varepsilon $}
			\State $u \gets u + 1$
		\EndWhile	
		\State $u_i \gets u-1$	
	\EndFor
	
	\Statex
	
	\For {$i \gets N-1$ to $0$}
	\While {$l \geq 0 \land \lvert x_i-x_l \rvert \leq \varepsilon $}
	\State $l \gets l - 1$
	\EndWhile	
	\State $l_i \gets l+1$	
	\EndFor
	
	\State \Return{$(L, U)$}
	\EndFunction
	\end{algorithmic}
\label{alg:calculate_neighborhood}

\end{algorithm}

\subsection{Neighborhood Size}

\label{sub:neighborhood_size}

Having the tables of lower and upper bounds $L=[l_0,...,l_{N-1}]$ and $U=[u_0,...,u_{N-1}]$ calculated, computing the size of $\varepsilon$-neighborhood of arbitrary point $x_i$ is simple. Recall that $N_\varepsilon(x_i)=\{x_{l_i}, ..., x_{u_i}\}$. We have $|N_\varepsilon(x_i)|=u_i - l_i +1$ which is trivial $O(1)$ operation.

The NeighborhoodSize algorithm \ref{alg:neighborhood_size} implements this calculation taking as parameters the index $i$ of a point $x_i$ and  bound tables $L$ and $U$.

\begin{algorithm}
	\caption{Calculating the size of the neighborhood}
	\begin{algorithmic}[1]
		\Require{$i \in \{0, ..., N-1\}$ point index}	
		\Require{$U=[u_0,...,u_{N-1}]$ upper index bounds for $X=[x_0, ...x_{N-1}] $} 
		\Require{$L=[l_0,...,l_{N-1}]$ lower index bounds for $X=[x_0, ...x_{N-1}] $} 
				
		\Statex
		\Function{NeighborhoodSize}{$i, L, U$}
		\State \Return{$u_i - l_i + 1$}
		\EndFunction
	\end{algorithmic}
	\label{alg:neighborhood_size}
\end{algorithm}

\subsection{Cluster Expansion}

\label{sub:cluster_expanstion}

In a single dimension, for the set of points $X=[x_1, ..., x_N]$ a cluster  can be described by a pair of its lower and upper bound indexes $(l, u)$.

By lemmas \ref{lemma:core_to_cluster} and \ref{lemma:cluster_by_core} we can find the cluster taking an arbitrary core point $p$ and points that are density reachable from $p$. From definition \ref{def:density_reachable} of density reachability we need to include points for which there is a chain such that consecutive points are directly density reachable. This in turn means that chain consecutive points are within $\varepsilon$-neighborhoods and earlier point in the chain is a core point.

We start from an arbitrary core point $p$ and its $\varepsilon$-neighborhood bounds $l_p$ and $u_p$ as cluster bounds $l$ and $u$. For upper cluster bound we iterate up through those temporary cluster points expanding the cluster $u$ bound if necessary. This action corresponds to following directly density reachable relation and later density reachable relation. We finish when reaching the moving target upper bound $u$ or the last point in dataset $X$. The case for the lower cluster bound $l$ is symmetric.

The ExpandCluster algorithm \ref{alg:expand_cluster} implements this idea. It takes as input the set of sorted input points $X$, the index of core point $p$ for which we will expand the cluster, the bound tables $L$ and $U$, table of cluster membership $C$, identifier for expanded cluster $clusterId$ and parameter $minPoints$.

In lines 2-3 the core point $x_p$ is assigned $clusterId$ and we initialize the cluster bounds to $l_p$ and $u_p$. 

Lines 6-14 implement the upper cluster bound expansion. The loop in line 6 iterates through temporary cluster points that can affect the cluster bound. If a point has not been visited yet or is marked as noise, it is assigned $clusterId$.  If we encounter a core point (line 9) the upper cluster bound is updated to its upper $\varepsilon$-neighborhood bound. The search is finished when we reach the last point in $X$ or the moving target upper cluster bound $u$. The case for lower cluster bound expansion in lines 16-24 is symmetric.

In line 26 we return new cluster description with the bounds $l$, $u$ and $clusterId$.

We touch every point in the cluster at most once. The complexity of the ExpandCluster algorithm \ref{alg:expand_cluster} is linear in the size of the cluster.

\begin{algorithm}
	\caption{Expanding the cluster for arbitrary core point}
	\begin{algorithmic}[1]
		\Require{$X=[x_0,...,x_{N-1}]$ sorted input points, $x_0 \leq ... \leq x_{N-1}$}
		\Require{$p \in \{0, ..., N-1\}$ core point index}	
		\Require{$U=[u_0,...,u_{N-1}]$ upper index bounds for $X=[x_0, ...x_{N-1}] $} 
		\Require{$L=[l_0,...,l_{N-1}]$ lower index bounds for $X=[x_0, ...x_{N-1}] $} 
		\Require{$C = [c_0, ..., c_{N-1}]$ cluster membership for $X=[x_0, ...x_{N-1}] $}
		\Require{$clusterId>0$ new cluster identifier}
		\Require{$minPoints > 0$}

		\Statex
		\Function{ExpandCluster}{$X, p, L, U, C, clusterId, minPts$}
		\State {$c_p \gets clusterId$}
		\State {$u \gets u_p$}
		\State {$l \gets l_p$}
		\State {}
		
		\For {$i \gets p+1$ to $N-1$ $\land$ $i \leq u$}
			\If {$c_i = NOT\_VISITED$}
				\State $c_i \gets clusterId$
					\If {\Call{NeighborhoodSize}{$i, L, U$} $ \geq minPts$}
						\State $u \gets u_i$
					\EndIf
			\ElsIf {$c_i = NOISE$}
				\State $c_i \gets clusterId$
			\EndIf
		\EndFor
			
		\State {}	
		\State $u \gets i - 1$
		\State {}	
		
		\For {$i \gets p-1$ to $0$ $\land$ $i \geq l$}
		\If {$c_i = NOT\_VISITED$}
		\State $c_i \gets clusterId$
		\If {\Call{NeighborhoodSize}{$i, L, U$} $ \geq minPts$}
		\State $l \gets l_i$
		\EndIf
		\ElsIf {$c_i = NOISE$}
		\State $c_i \gets clusterId$
		\EndIf
		\EndFor

		\State {}	
		\State $l \gets  i + 1$
		\State {}	
					
		\State \Return{\Call{NewCluster}{$l,u, clusterId$}}
						
		\EndFunction
	\end{algorithmic}
	\label{alg:expand_cluster}
\end{algorithm}

\subsection{DBSCAN1D Complexity}

Now that we examined all the sub-functions of DBSCAN1D algorithm \ref{alg:dbscan1d} we can analyze the complexity of the whole algorithm.

Initialization part in lines 2-6 takes $O(N)$ time. The CalculateNeighborhood algorithm \ref{alg:calculate_neighborhood} also needs $O(N)$ time as can be seen in section \ref{sub:calculating_neighborhood}.

The loop in lines 8-16 iterates through all the $N$ input points in $X$. A cluster is expanded only if the point has not been visited yet and it is a core point. From section \ref{sub:cluster_expanstion} we know that the ExpandCluster algorithm \ref{alg:expand_cluster} runs in time linear in the size of cluster and marks all the points as visited with $clusterId$. The core point check executed in line 11 with NeighborhoodSize algorithm \ref{alg:neighborhood_size} is $O(1)$ operation discussed in section in \ref{sub:neighborhood_size}.

Summing up, if $C_1, ... C_k$ are the clusters of the set $X$ and $noise$ is the set of points that don't belong to any cluster the overall complexity of the algorithm is $O\left(\sum_{i=1}^{k}|C_i| + |noise|\right)$. By definition we have $N=\sum_{i=1}^{k}|C_i| + |noise|$ which means that DBSCAN1D algorithm \ref{alg:dbscan1d} runs in $O(N)$ time. It is worth nothing that the complexity does not depend on $\varepsilon$ and $minPoints$ parameters.

The space complexity of the algorithm is also $O(N)$. The algorithm only uses auxiliary tables of size $N$ and a list of the clusters.

The formulation of the algorithm assumed that the table with input points $X$ is sorted. If this is not the case one needs additional sorting step before calling the algorithm which in general case takes $O(NlogN)$ time.

\subsection{Implementation Notes}

\label{sub:implementation_notes}

If the algorithm is called repeatedly for input of known or bounded size $N$ the memory can be allocated just once in advance. 

When the algorithm returns with a list of clusters the order of points within each cluster is no longer relevant. This means that for each cluster separately we can resort with different order and recursively call DBSCAN1D in place. Similarly one can re-cluster with different parameters $minPoints$ and $\varepsilon$. In this case resorting step is not necessary.

The original paper \cite{original} assigns border points only to the first cluster which expands into them. If there is need to assign border points to all the clusters they belong to, it is enough to modify ExpandCluster algorithm \ref{alg:expand_cluster} to include point into cluster even if it was already visited. Alternatively border points can be marked as noise as in \cite{border_noise}.

The algorithm can be modified to work with wrap-around data. As an example we can consider angular data where $x_i \in [0, 2\pi)$. Where distances between points are concerned the data wraps around $2\pi$, e.g. the distance between $x_1=0$ and $x_2=3/2\pi$ is $\pi/2$. The distance calculation has to take into account the above mentioned circular nature but additionally neighborhood calculation and cluster expansion have to wrap around the input data $X$ with modular arithmetic. As an example consider $\varepsilon=pi/2$ and $X=[0, pi/4, \pi, 2\pi - pi/4 ]$. The $\varepsilon$-neighborhood of $x_3=2\pi - \pi/4$ will contain $x_1=\pi/4$ and $x_3$ upper bound index $u_3=N+1=5$. In a symmetric manner $x_1=\pi/4$ lower bound index $l_1=-1$ which points to $x_3=2\pi - \pi/4$ entry in $X$. The tables $X$, $L$, $U$ and cluster descriptions have to work in modular arithmetic and handle indexes with greater than N or negative values.

\subsection{Experimental Results}

A simple R \cite{R} package backed by Rcpp \cite{rcpp2, rcpp1} C++ implementation was written. There are two available R packages for comparison on CRAN. The \emph{fpc} \cite{fpc} DBSCAN is a straightforward $O(N^2)$ R implementation. The \emph{dbscan} \cite{dbscan} package has optimized C++ implementation that uses k-d trees for neighbor search. The \emph{fpc} package was ruled out from comparison after preliminary tests as it could not compete with non-naive C++ implementations.

In all the experiments in this section the sorting time is included in DBSCAN1D running time effectively making it $O(NlogN)$ complexity algorithm.

\begin{figure}[h] 
	\centering
	\includegraphics[scale=0.5]{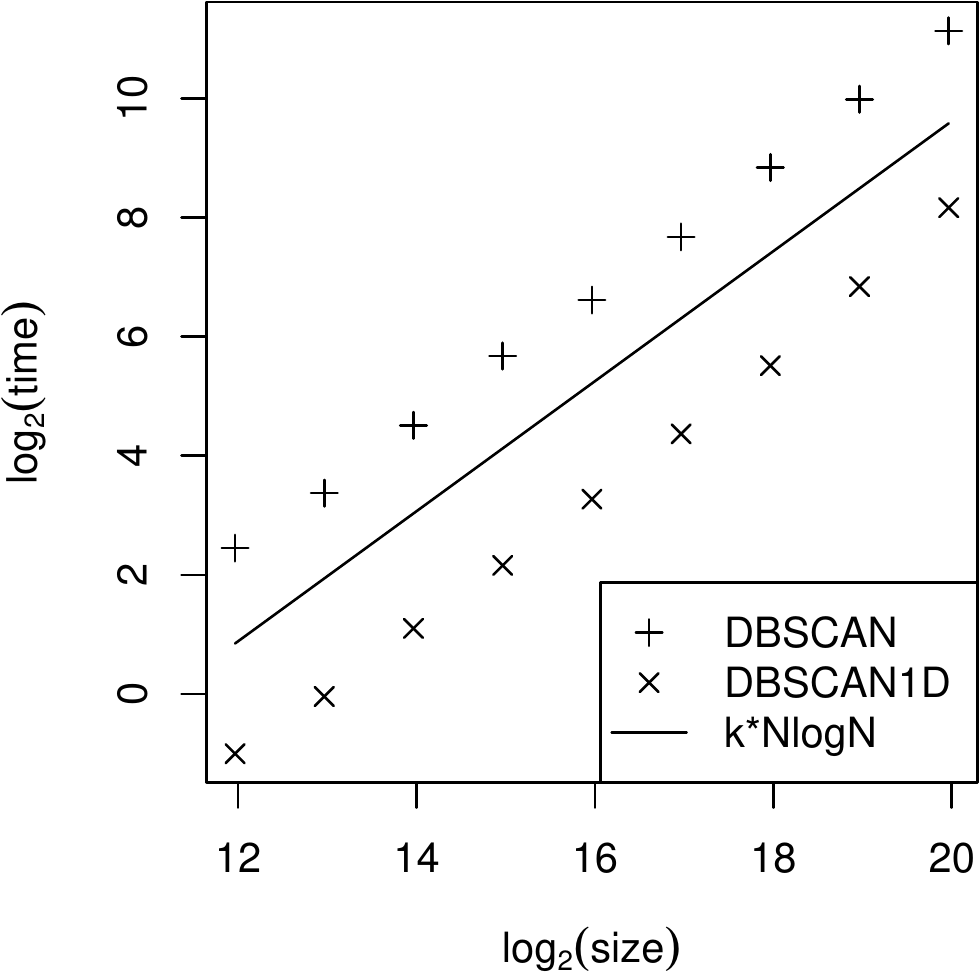}
	\caption{Execution time dependence on input size when $|N_\varepsilon(x_i)|\leq logN$. $NlogN$ multiplied by constant was plotted as reference. }
	\label{fig:execution_vs_input}  
\end{figure}

The first experiment is setup to benchmark execution time dependence on input size when $\varepsilon$-neighborhoods are small, $|N_\varepsilon(x)|\leq logN$ on average. A number of separated clusters were generated from uniform distribution. For exponentially increasing input size two algorithms were called 10 times and mean execution time was taken. The log-log plot in figure \ref{fig:execution_vs_input} shows that algorithms from \emph{dbscan} package and this paper run in $O(NlogN)$ time when $|N_\varepsilon(x)|\leq logN$. As a side note DBSCAN1D was faster by a constant factor.

\begin{figure}[h] 
	\centering
	\includegraphics[scale=0.5]{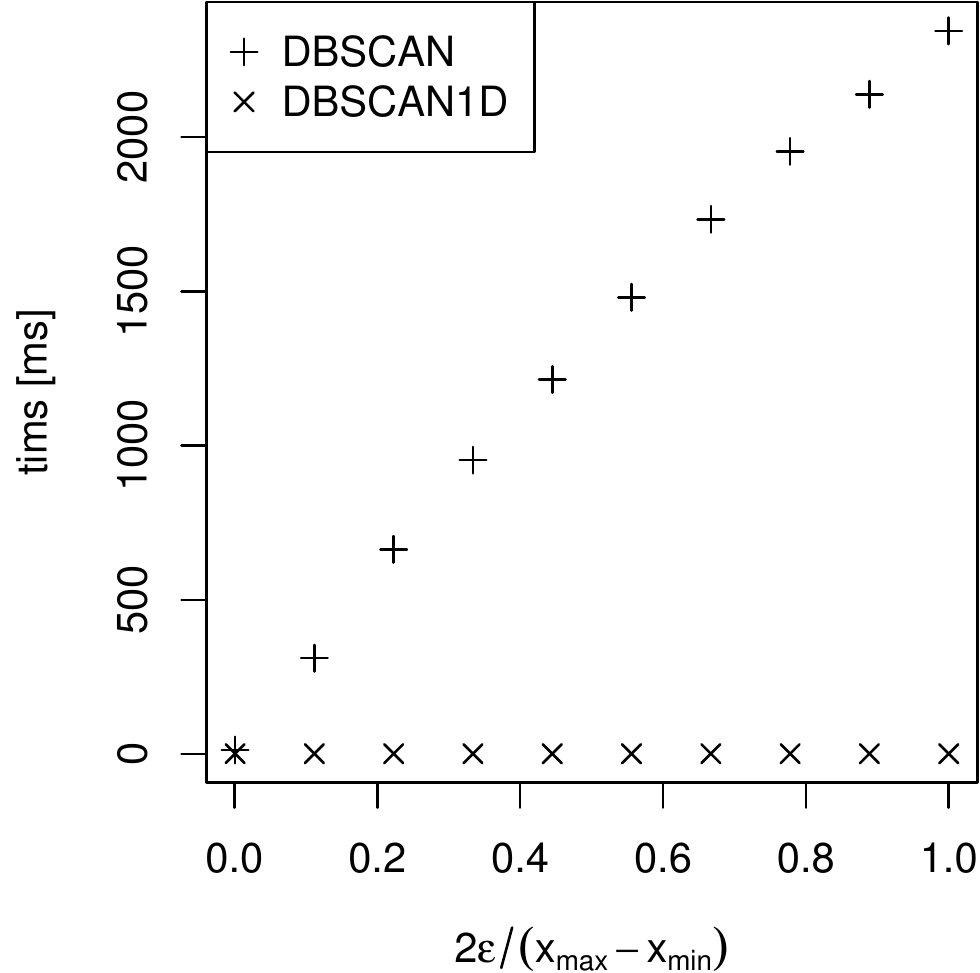}
	\caption{Execution time dependence on $\epsilon$. DBSCAN1D running time does not depend on $\varepsilon$ value.}
	\label{fig:execution_vs_eps}  
\end{figure}

The second experiment is setup to benchmark execution time dependence on $\varepsilon$ size. Input size was fixed and $\varepsilon$ was varied. Figure \ref{fig:execution_vs_eps} confirms that DBSCAN1D running time is not dependent on $\varepsilon$ value. The implementation from \emph{dbscan} package suffers from the original DBSCAN problem described in \cite{revisited, faster}. The running time approaches the pessimistic $O(N^2)$ complexity as $\varepsilon$ and $\varepsilon$-neighborhood sizes are increasing. 

The DBSCAN1D R package is available on github \cite{dbscan1d}. Benchmark results can be recreated through package vignette.

\section{Application}

This section focuses on DBSCAN1D algorithm \ref{alg:dbscan1d} applied to line extraction from 2D laser range data. 

The problem arises in feature based robotics localization, mapping and SLAM. An overview and experimental evaluation of six popular algorithms used in mobile robotics and computer vision can be found in \cite{line_extraction}.

\subsection{Preliminaries}

\subsubsection{2D Laser Range Data Segmentation}

\begin{figure}[h]
	\centering
	{\includegraphics[scale=0.5]{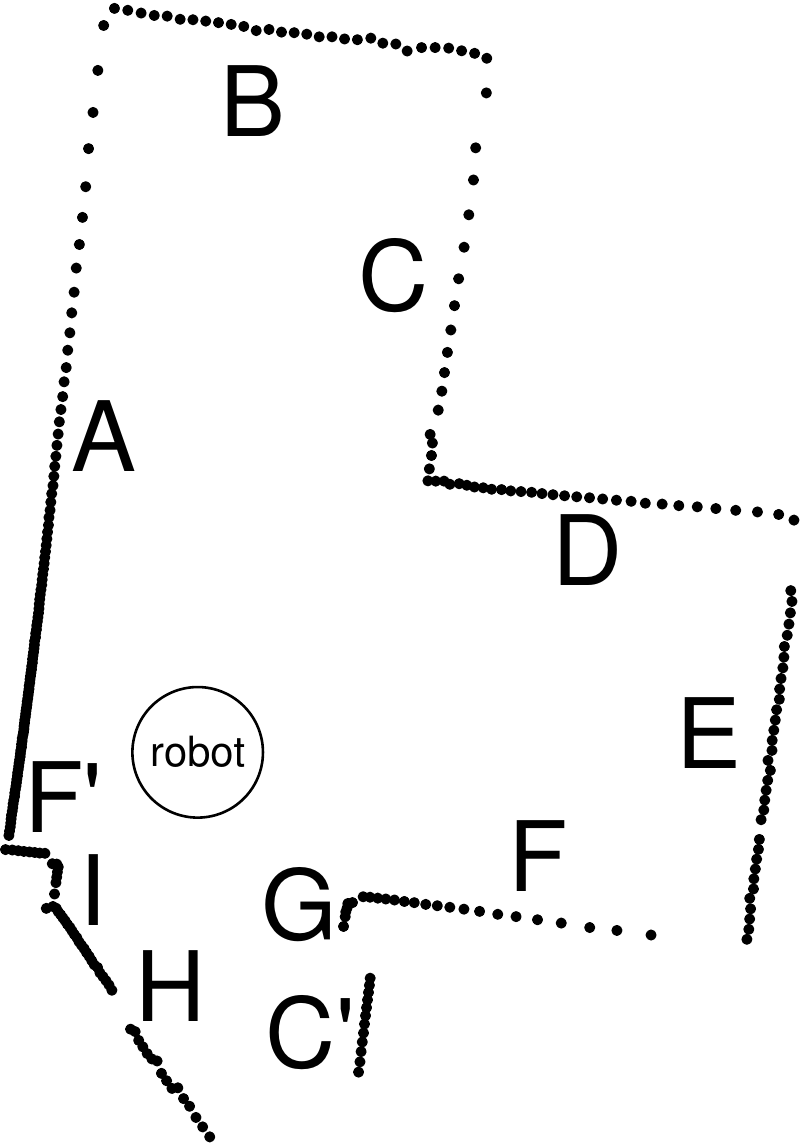}}
	\caption{A robot equipped with lidar retrieves 360 degree scan. The task is to segment the scan points into groups forming linear features (marked by capital letters). The points that can't be grouped this way should be marked as noise.}
	\label{fig:segmentation}
\end{figure}

A robot scans its surroundings with laser. In effect we get a point cloud lying on the plane. The problem is to extract features from the point cloud that can be of use for further processing. In this paper we focus on extracting line features as seen in figure \ref{fig:segmentation}.

\subsubsection{Line Representation}

\label{subsub:line_representation}

A common slope-intercept representation of a line $y=mx+b$ is not suitable for computational geometry \cite{pdim}. The slope coefficient $m$, which has value of tangent, grows unbounded for nearly vertical lines.  

\begin{figure}[h]
	\centering
	{\includegraphics[scale=0.5]{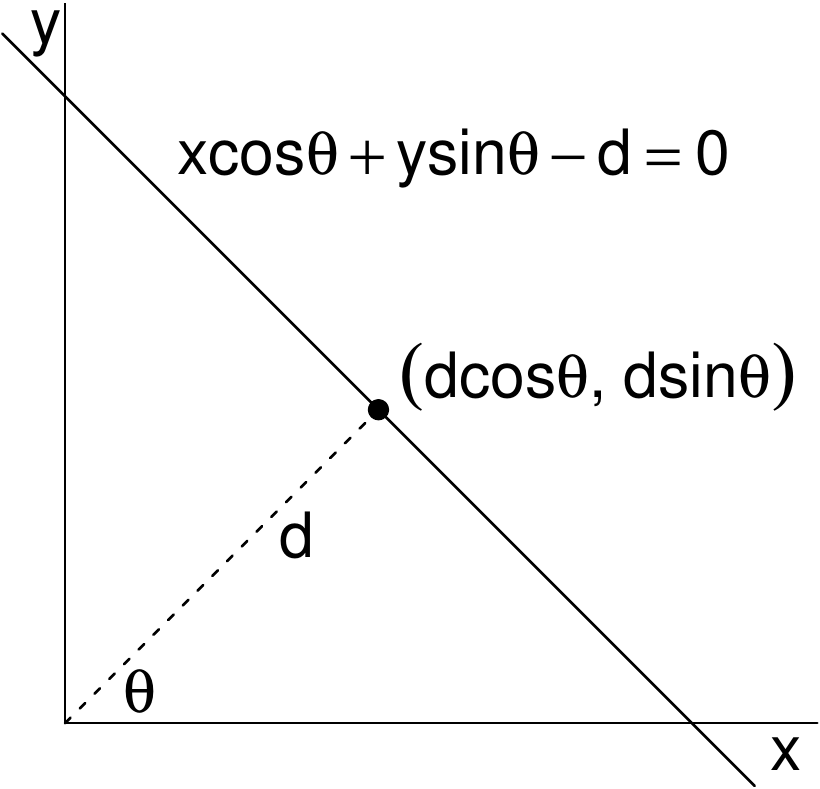}}
	\caption{A normal form representation of a line in two dimensional Cartesian space. Here $d$ is the distance of the line to the origin and $\phi$ is the oriented angle between unit vector of the x-axis and segment joining origin with line's nearest point. For computational purposes line is defined by a pair $(d, \phi)$.}
	\label{fig:normal_form}
\end{figure}

General form representation $Ax + By + C = 0$ does not suffer from the above problem. One can normalize the coefficients by dividing the $A$, $B$, $C$ by ${|C| \over -C}  \sqrt{A^2 + B^2}$. After normalization one gets normal form representation of a line $x cos \theta + y sin \theta -d = 0$. In the normal form line is represented by polar descriptors $(d, \theta)$. The geometric interpretation of polar descriptors is shown in figure \ref{fig:normal_form}. The coefficient $d$ is the distance of the line from the origin. The coefficient $\theta$ is oriented angle between unit vector of the x-axis and segment from the origin perpendicular to the line. In other words $(d, \theta)$ are polar coordinates of the point on line closest to the origin.

The normal form of the line has one more property that shall proof useful for us. The distance from arbitrary point $(x,y)$ to the line is given by $r=|x cos \theta + y sin \theta -d|$. The value $x cos \theta + y sin \theta -d$ has positive sign if $(x,y)$ and $(0,0)$ lie on the distinct sides of the line, negative if they lie on the same side and zero if the point lies on the line.

\subsubsection{Line Fitting}

\label{subsub:line_fitting}

Ordinary least squares (OLS) regression method is not good choice for problems where the errors are on both $X$ and $Y$ coordinates. The method assumes that one of the coordinates is known without error.

For geometric problems better results can be obtained with total least squares (TLS) method which minimizes the perpendicular distances from the points to the line.

TLS formulas for normal line representation can be found in \cite{arras} including weighted case and covariance matrix calculation. For the unweighted case, assuming data $(x_0,y_0), ..., (x_{N-1}, y_{N-1})$, $\bar{x}=1/N\sum_i x_i$ and $\bar{y}=1/N\sum_i y_i$ the solution is given by: 

\begin{equation*}
tan 2 \theta = {-2 \sum_i(\bar{y} - y_i)(\bar{x} - x_i )  \over \sum_i \left[ {\left( \bar{y} - y_i \right)}^2 - {\left( \bar{x} - x_i \right)}^2 \right]}
\end{equation*}

\begin{equation*}
d = \bar{x}  cos\theta + \bar{y}  sin\theta
\end{equation*}

After calculating the arctangent function one usually normalizes the result for correct $\theta$ range and $d$ sign.

\subsubsection{Circular Mean}

\label{subsub:circular_mean}

The arithmetic mean is not suitable for calculating average value of angle values with direction interpretation. The problem arises when we wrap around $2\pi$. As an example consider average value for $\pi/2$ and $3/2\pi$. The arithmetic mean incorrectly yields $pi$. The correct solution should be $0$.

The remedy for this problem \cite{circular_statistics} is to convert each angle $\alpha_i$ to corresponding points on unit circle $(cos \alpha_i, sin\alpha_i)$, compute the arithmetic mean over coordinates and convert back the resulting point to polar representation with arctangent function.

The circular mean is given by the formula $\bar{\alpha}=atan2\left(\sum_i sin\alpha_i, \sum_i \cos \alpha_i  \right)$ where \emph{atan2} is a variant of arctangent function commonly used in computational geometry. 

\subsection{Algorithm Overview}

When retrieving scan point data from hardware there is additional implicit information in the order of points. Using this data we may estimate the local angle at each point. Scan points corresponding to linear features form high density regions in local angle space. The algorithm will cluster the data in angular space.

After angular segmentation one gets clusters of points that share local angle. Points may belong to lines, parallel lines or should be considered as noise. Parallel lines can be separated based on distance to any arbitrary line that shares the same angle. We estimate the angle as cluster mean and use line through the origin to separate distinct parallel lines.

The AngularSegmentation algorithm \ref{alg:segmentation} implements this idea. It takes as input scan points data and clustering parameters $\varepsilon_\theta$, $\varepsilon_{dist}$ and $minPoints$. In lines 4-6 we perform segmentation in local angle space. In lines 8-13 we consider each cluster with parallel lines separately. We segment based on distance to the line through origin with cluster angle. In line 15 we return the found clusters.

AngularSegmentation shares some ideas with classic Hough Transform \cite{computer_vision} computer vision method for line extraction. In Hough Transform each point votes for all lines that would pass through it in line polar descriptors $(d, \theta)$ discretized space. The method suffers from two problems as described in \cite{computer_vision}. It is difficult to choose appropriate size for discretization grid and method generates non-existing lines with noisy data. AngularSegmentation does not depend on grid discretization, has build in mechanism for handling noise and is computationally efficient.

\begin{algorithm}
	\caption{Laser range finder segmentation}
	\begin{algorithmic}[1]
		\Require{$scan$ data with $X=[x_0, x_{N-1}]$, $Y=[y_0, y_{N-1}]$}, empty $\theta$ and $dist$	
		\Require{$\varepsilon_\theta>0, \varepsilon_{dist}>0, minPoints>0$ clustering parameters}		
		\Statex
		\Function{AngularSegmentation}{$scan, \varepsilon_\theta, \varepsilon_{dist}, minPoints $}
			\State $finalClusters \gets \emptyset$

			\State 
		
			\State $scan.\theta \gets$  \Call{EstimateLocalAngle}{scan}
			\State $scan.\theta \gets$ \Call{Sort}{$scan_\theta$}
			\State $clusters_\theta \gets$ \Call{Dbscan1D}{$scan.\theta$, $\varepsilon_\theta$, $minPoints$} 
		
			\State 
			
			\ForAll{C in $clusters_\theta$}
				\State $mean_\theta \gets$ \Call{CircularMean}{$scan.\theta[C]$}
				\State $scan.dist[C] \gets$ \Call{PointLineDistance}{$scan.X[C], scan.Y[C], mean_\theta$}
				\State $scan.dist[C] \gets$ \Call{Sort}{$scan.dist[C]$}
				\State $cluster \gets$ \Call{Dbscan1D}{$scan.dist[C]$, $\varepsilon_{dist}$, $minPoints$}
				\State $finalClusters \gets finalClustres \cup cluster $
			\EndFor
		
			\State
		
			\State \Return{$finalClusters$}
		\EndFunction
	\end{algorithmic}
	\label{alg:segmentation}
\end{algorithm}

\subsection{Local Angle Estimation}

For local angle estimation total least squares (TLS) method described in section \ref{subsub:line_fitting} was used. TLS is run for all the triplets of consecutive points in a loop. We are only interested in line angle $\theta$ and the $d$ parameter does not need to be calculated.

There are other possible ways to estimate local angle but this is beyond the scope of this work.

\subsection{Angular Segmentation}

After estimating the local angle we sort the scan points by local angle. Both of these operations can be performed while collecting the scan points from hardware if time is critical.

In the next step we call DBSCAN1D algorithm \ref{alg:dbscan1d} adapted for circular data as described in section \ref{sub:implementation_notes}. We obtain a set of clusters where each cluster groups points lying on various parallel lines.

\subsection{Parallel Lines Segmentation}

Each angular cluster may group points from distinct parallel lines that share local angle. We will consider each such cluster separately. The points from distinct parallel lines can be separated by distance to arbitrarily chosen parallel line. The simplest choice is the line going through the origin.

The line through the origin has polar descriptors $(0, \theta)$ and equation $x cos \theta + y sin \theta = 0$ as described in section \ref{subsub:line_representation}. A good estimate of line angle $\theta$ is circular mean of cluster angles calculated as described in section \ref{subsub:circular_mean}.

Once we know the angle $\theta$ for line going through the origin we can calculate the distance to all the points in the cluster. As we discussed in section \ref{subsub:line_representation}, for normal line form, the distance can be calculated as $r=|x cos \theta + y sin \theta|$ or $x cos \theta + y sin \theta$ if one is interested in point, origin, line relation encoded in the sign. We use the later formula.

Having calculated the distances from cluster points to line through origin we proceed in segmentation. We sort the scan sub-array that corresponds to the cluster and call DBSCAN1D algorithm \ref{alg:dbscan1d} for distance data. The found clusters are added to final solution and we proceed to next angular cluster.

\subsection{Collinear Lines Separation}

The clusters after parallel lines segmentation may contain points from various collinear lines that are not continuous. One can add another separation layer but this is beyond the scope of this work.

\subsection{AngularSegmentation Complexity}

As far as complexity is concerned the most time consuming operation in the algorithm \ref{alg:segmentation} are $O(NlogN)$ sort operations in lines 5 and 11. Local angle estimation, DBSCAN1D, circular mean and point-line distance computations are all linear in their input size.

\subsection{Implementation Notes}

The local angle estimation and sorting from lines 4-5 can be executed while collecting data from the hardware.

The memory for DBSCAN1D algorithm \ref{alg:dbscan1d} can be reserved in advance. The number of points returned by the hardware in each scan is known a priori.

While estimating local angle in line 4  we iterate over triplets in scan data. One can keep partial TLS sums and update them during iteration. While this is tempting, care has to be taken for accumulated floating point error.

The computations for angle clusters in lines 8-13 can be performed in place on original scan data as discussed in section \ref{sub:implementation_notes}.

The local angle estimate from line 4 was normalized to range $[0, \pi)$. This way parallel features on distinct sides of the robot, like walls, strengthen themselves together.

Some of the collected readings are marked as faulty by hardware. For triangulation lasers the reasons may include too short or long distance. Such readings are ignored in the algorithmm.

\subsection{Shortcomings}

If features are very close to each other, either in angular space with near angles or in distance space, there is a risk that they will be clustered together. It is possible to rerun segmentation for single cluster with more conservative $\varepsilon$-neighborhood but the problem lies in detecting those situations reliably. 

In theory arcs points observed from close enough distance can be clustered together. The points will have very close local angle estimates. Adding a layer to the algorithm that clusters based on local curvature measure could solve the problem. 

\subsection{Experimental Implementation}

Open source algorithm implementation is available in ev3dev-mapping-ui \cite{ev3dev-mapping-ui}. Preliminary experiments show that algorithm runs in millisecond order time and is resistant to noise.

A video describing algorithm with real-time visualization is available \cite{angular-video} online.

\section{Conclusion}

Initial experiments show that AngularSegmentation algorithm \ref{alg:segmentation} is capable of working in real time and robust to noise. It may be useful in practice. A detailed study of the algorithm performance is necessary. Study should include objective benchmark methods and comparison with state of the art algorithms, similar to the one performed in \cite{line_extraction}. 

If algorithm proofs useful compared to competitors, efficient implementation for plane extraction from 3D range data should be evaluated. Such implementation could take advantage of low complexity DBSCAN algorithms for higher dimensions as in \cite{revisited, revisited_long, faster}.

Other research directions may include extraction of non-linear features, re-segmentation of merged features and evaluation of various local angle estimation methods.

\section{Acknowledgements}

The idea emerged around 2016 when I was working at National Centre for Nuclear Research Świerk. At this time, I contributed to ev3dev open source OS for Lego Mindstorms EV3 \cite{ev3dev}, the initiative started by Ralph Hempel and David Lechner.
As one of my hobby projects I interfaced Neato XV-11 lidar to Lego Mindstorms EV3 \cite{ev3dev-xv11}. This sparked the idea for segmentation algorithm.

Back in Świerk, with my friends - Marcin Buczek, Mariusz Śmierzyński, Michał Andrasiak and Agnieszka Misiarz, we had the habit of discussing all the ideas, no matter how funny, crazy or stupid. Something I miss to this day. Numerous times I brought the idea for this algorithm to discussion. This is also the time when I implemented rough prototype of the algorithm in R and started writing technical description.

In 2017 I began my work at Industrial Research Institute for Automation and Measurements PIAP, now Łukasiewicz Research Network - Industrial Research Institute for Automation and Measurements PIAP. As part of the contract negotiations I was allowed to finish the technical description during some time of the first month of my work.

Several years later in 2021, I found the technical description on my laptop. This is the paper you are reading now.

\printbibliography

\end{document}